\documentclass[letterpaper]{article} 
\usepackage{aaai24}  
\usepackage{times}  
\usepackage{helvet}  
\usepackage{courier}  
\usepackage[hyphens]{url}  
\usepackage{graphicx} 
\usepackage{natbib}  
\usepackage{caption} 
\usepackage{algorithm}
\usepackage{algorithmic}

\usepackage{newfloat}
\usepackage{listings}
\DeclareCaptionStyle{ruled}{labelfont=normalfont,labelsep=colon,strut=off} 
\floatstyle{ruled}
\newfloat{listing}{tb}{lst}{}
\floatname{listing}{Listing}
\usepackage[table,xcdraw]{xcolor}
\usepackage{booktabs}
\usepackage{mathrsfs}
\usepackage{multirow}
\usepackage{subfigure}
\usepackage{amssymb}
\usepackage{amsmath}
\usepackage{pifont}

\usepackage{amsmath,amsfonts,bm}

\def\eqref#1{equation~\ref{#1}}

\def\1{\bm{1}}

\def\vx{{\bm{x}}}

\def\vz{{\bm{z}}}

\DeclareMathAlphabet{\mathsfit}{\encodingdefault}{\sfdefault}{m}{sl}
\SetMathAlphabet{\mathsfit}{bold}{\encodingdefault}{\sfdefault}{bx}{n}

\newcommand{\app}{\raise.17ex\hbox{$\scriptstyle\sim$}}

\definecolor{deemph}{gray}{0.6}

\definecolor{baselinecolor}{gray}{.9}

\newcommand{\cmark}{\ding{51}}%
\newcommand{\xmark}{\ding{55}}%
\newcommand{\breakrows}[1]{\begin{tabular}{c} #1 \end{tabular} }

\title{
Masked Momentum Contrastive Learning for Zero-shot Semantic Understanding
}
\vspace{-1em}
\author{
    Jiantao Wu$^1$\thanks{jiantao.wu@surrey.ac.uk}, Shentong Mo$^2$, Muhammad Awais$^1$, Sara Atito$^1$, \\ Zhenhua Feng$^1$, Josef Kittler$^1$
}
\affiliations{
$^1$ University of Surrey, $^2$ Carnegie Mellon University
}

\begin{document}

\maketitle

\vspace{-1em}

%


\begin{abstract}
Self-supervised pretraining (SSP) has emerged as a popular technique in machine learning, enabling the extraction of meaningful feature representations without labelled data. In the realm of computer vision, pretrained vision transformers (ViTs) have played a pivotal role in advancing transfer learning. Nonetheless, the escalating cost of finetuning these large models has posed a challenge due to the explosion of model size. This study endeavours to evaluate the effectiveness of pure self-supervised learning (SSL) techniques in computer vision tasks, obviating the need for finetuning, with the intention of emulating human-like capabilities in generalisation and recognition of unseen objects. To this end, we propose an evaluation protocol for zero-shot segmentation based on a prompting patch. Given a point on the target object as a prompt, the algorithm calculates the similarity map between the selected patch and other patches, upon that, a simple thresholding is applied to segment the target. Another evaluation is intra-object and inter-object similarity to gauge discriminatory ability of SSP ViTs. Insights from zero-shot segmentation from prompting and discriminatory abilities of SSP led to the design of a simple SSP approach, termed MMC. This approaches combines \textbf{M}asked image modelling for encouraging similarity of local features, \textbf{M}omentum based self-distillation for transferring semantics from global to local features, and global \textbf{C}ontrast for promoting semantics of global features, to enhance discriminative representations of SSP ViTs. Consequently, our proposed method significantly reduces the overlap of intra-object and inter-object similarities, thereby facilitating effective object segmentation within an image. Our experiments reveal that MMC delivers top-tier results in zero-shot semantic segmentation across various datasets. Moreover, it offers competitive performance for finetuning and in scenarios with limited data, highlighting its effectiveness in capturing semantic nuances.

\end{abstract}

\section{Introduction}
The self-supervised pretrained (SSP) transformers have significantly influenced the fields of natural language processing (NLP)~\cite{brown2020language} and computer vision (CV)~\cite{atito_sit_2021}, chiefly due to the ability to capture high-level feature representations without the relying on labeled data. In NLP, the dominant method for pretraining is masked language modelling (MLM)~\cite{devlin2018bert,brown2020language}. While in CV, there are two main streams of self-supervised learning (SSL) methods, namely, view-invariance methods which work with global embeddings such as \cite{chen_simple_2020,grill2020bootstrap,caron_emerging_2021,mo2021spcl,mo2022pauc} and masked image modelling (MIM) based methods, such as \cite{atito_sit_2021,bao_beit_2022,xie_simmim_2022,he_masked_2022}. Self-supervised pretraining of DNNs on large datasets can attain improved performance when finetuned on various downstream tasks, i.e. the target problems, predominantly outperforming supervised pretraining~\cite{DBLP:conf/nips/ChenKSNH20}. 

However, the finetuning process is increasingly becoming cost-prohibitive due to the escalating size of the models, e.g. GPT 4 has over one trillion parameters~\cite{schreiner2023gpt4}. In NLP, large language models (LLMs) circumvent this issue due to their zero/few-shot capabilities from prompts, i.e. zero/few-shot learning without any gradient updates or finetuning~\cite{brown2020language}. 
Adopting the zero-shot prompting with SSL paradigm, familiar within NLP, is not commonplace in the field of CV. 
Motivated by LLMs, we aim to establish an evaluation protocol for assessing the ability of SSP ViT \cite{dosovitskiy2020vit} for zero-shot prompting instead of just finetuning or linear evaluation via gradient updates. Compared to $k$-NN protocol, our setting is more natural in that humans do not need some irrelative samples to help them recognise objects.

A significant hurdle for zero/few-shot prompting lies in designing an appropriate prompt that instructs the model regarding the task at hand. In NLP, words or semantic instructions are readily provided through typing, e.g., instructing ChatGPT to translate by  ``translate: \{text\}''. However, obtaining a semantic instruction from a user in the context of computer vision poses a challenge, primarily due to the complex nature of visual data and the lack of a direct and structured interface as NLP.

\begin{figure}[t]
    \centering
    \includegraphics[width=.9\linewidth]{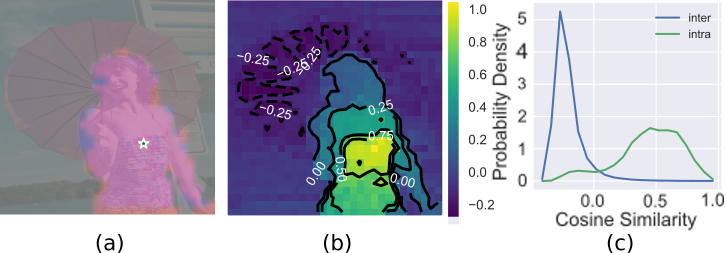}
    \caption[short]{A zero-shot segmentation example for segmenting the female. (a) Selected patch/click ($\bigstar$), ground-truth label (blue), and segmentation mask by our method (pink) which is almost covering the blue ground-truth mask. (b) Heatmap of similarities between the selected patch. (c) Distributions of intra-object and inter-object similarities.}\label{fig:front_features}
\end{figure}

In computer vision, many zero/few-shot learning tasks require language instructions to guide the behaviour of the models. For instance, UniBoost~\cite{DBLP:journals/UniBoost} finetunes unsupervised unimodal models to boost zero-shot vision-language (VL) tasks. However, vision-language pretraining (VLP) faces multimodality challenges that depend on human-generated image-text pairs, which incorporates weak supervision. 
Moreover, methods requiring VLP for downstream tasks may not be suitable to evaluate SSP models from vision only modality. 
To avoid language modality, some methods propose using interactive queries or prompts, such as point clicks, to segment the target object in an image, as done by SAM~\cite{sofiiuk2022reviving,kirillov2023segment}. However, the above methods require supervision or finetuning to acquire the ability of generalisation. We argue that SSP models have the potential on learning such ability unsupervisedly.

In this work, we explore the ability of SSP ViTs to learn visual representations without introducing extra knowledge or other modality. 
The standard protocols to evaluate vision based SSL is to use finetuning or linear evaluation for classification and segmentation. 
However, linear evaluation may favour invariance based methods, like contrastive learning which tend to learn the dominant concept~\cite{chen2021intriguing}. 
Hence, for a depth understanding of the representation capabilities of SSP ViTs, we include a zero-shot prompting evaluation protocol along with linear evaluation and finetuning. 
For this, we introduce a straightforward framework utilising the principle of threshold-based algorithm for segmentation, leveraging cosine similarity for decision-making. 
By requiring only one patch/click for segmentation as a prompt, the framework offers a more streamlined and potentially efficient approach for the evaluation of segmentation tasks. 
For example, we select a patch in an image and compute the cosine similarity map between representations of this patch and all other patches. We then filter regions above a certain threshold in the similarity map to obtain a segmentation mask as demonstrated in Figure~\ref{fig:front_features}. 

Our analysis shows that simple MIM methods struggle with zero-shot prompting due to high inter-object similarity and the self-distillation on local features can significantly enlarge the distinction between inter- and intra- objects.
In line with these insights, we found out that a simple mechanism to merge \textbf{M}asked image modeling (MIM), \textbf{M}asked momentum based self-distillation, and global \textbf{C}ontrastive learning can lead to good a SSL method, namely, MMC.
Specifically, we introduce contrastive learning (CL) on the CLS token to learn global semantic image representation, momentum-based self-distillation on patch tokens to learn local semantics and to transfer global semantics from CLS token to local patches, and MIM based reconstruction to learn internal semantic structures. 
This integration enhances intra-object similarity, while creating a noticeable gap between intra-object and inter-object similarities, and yields excellent performance on segmentation tasks while maintaining good performance on classification tasks.

We employ the proposed zero-shot prompting algorithm to evaluate popular SSL models, as well as the proposed MMC. 
We also analyse inter-object and intra-object similarity distributions to reveal the limitations of SSL models. 
Interestingly, methods that combine MIM with view-invariance priciples~\cite{atito_sit_2021,zhou_ibot_2022} can significantly encourage discrimination between intra- and inter- object similarity. 
The proposed MMC achieves comparable performance to state-of-the-art (SOTA) SSL approaches in classification finetuning settings on ImageNet-1K. However, MMC outperformed existing SSL approaches with large margins for zero-shot and linear evaluation for segmentation. 
Based on the proposed framework and model, we develop an online demo to showcase the capability of MMC for zero-shot segmentation at \url{https://huggingface.co/spaces/erow/SimilarityThresholding}.

The main contributions of this paper include: a) The evaluation protocol and insights for SSL for zero-shot segmentation from prompting and discriminatory analysis via inter-object and intra-object similarity. 
b) We empirically find that 1) Reconstruction pretext increases both intra-object as well as inter-object similarities. 2) Momentum distillation on the local patches is effective to encourage the discriminative local features for objects' separation and distillation of global discriminative semantics to local features.
c) Based on these insights we develop a simple SSL approach that utilises MIM, momentum distillation, and global contrast to capitalise local and global semantics with discriminatory abilities. 
d) The proposed method exhibits promising results across various segmentation tasks, notably in zero-shot segmentation with prompting, while also demonstrating competitive performance in classification and transfer learning for finetuning.
 
\section{Related Work}

\subsection{Self-supervised Learning}

\noindent\textbf{Masked Image Modeling (MIM).} is a popular self-supervised learning brunch in computer vision. It is motivated by the masked language modeling in NLP~\cite{vaswani2017attention,devlin2018bert} and requires models to predict invisible parts of input data from a masked view.
Initial attempts in MIM, like iGPT~\cite{chen2020generative} and ViT~\cite{dosovitskiy2020vit}, do not surpass the supervised ones until SiT~\cite{atito_sit_2021}.
Since then, researchers have studied masking targets, including HOG features~\cite{wei_masked_2021}, visual tokens from a pretrained discrete VAE~\cite{ramesh2021zero}, and so on~\cite{he_masked_2022,xie_simmim_2022,chen_context_2022,wu2022objectwise}.
SiT~\cite{atito_gmml_2022} reveals that a simple pixel reconstruction task with a high masking proportion is a strong visual representation learner.

\noindent\textbf{View-invariance Learning.}\quad
Contrastive learning (CL) is a popular technique to encourage invariance between differently augmented views of the same image, such as MoCo~\cite{he_momentum_2020}, SimCLR~\cite{chen_simple_2020}, and Barlow Twins~\cite{bardes2022vicregl,zbontar_barlow_2021}. MoCo v3~\cite{chen_empirical_2021} uses a momentum encoder to learn consistent representations on CLS token with ViTs to achieve SOTA performance. 
VicRegL~\cite{bardes2022vicregl} applies the VICReg criterion~\cite{bardes2021vicreg} to global and local features for learning invariant and semantic-aware representations.
\citeauthor{caron_emerging_2021} shows that self-supervised ViTs are more efficient to learn explicit information about the semantic segmentation of an image than supervised models through self-distillation.

\noindent\textbf{Masking-invariance Learning.}\quad{}
Recent progress in masking-invariance techniques has garnered scholarly interest~\cite{atito_sit_2021,assran2022masked_MSN}. These methods, inspried by MIM, dividing an image into a masked view and its unmasked view, aim to enhance the congruence of representations from two views~\cite{yao2022masked,huang2022contrastive,tao2022siamese}. 
Specifically, MSN~\cite{assran2022masked_MSN} employs anchor views with masked patches to align with the target views with whole patches by optimizing the cross-entropy of the anchor and the target, which performs competitively on low-shot image classification. 
SiT~\cite{atito_sit_2021}, combining MIM and masked contrast on the CLS token, outperforms with a large margin on many small datasets.
MC-SSL~\cite{atito_mc-ssl00_2021}, combining MIM and masked contrast on patch tokens, demonstrates strong concept information on patch tokens.
Inspired by DINO~\cite{caron_emerging_2021} and MIM methods, iBOT~\cite{zhou_ibot_2022} introduces self-distillation on masked patch tokens, as well as the CLS token, to borrow the global semantics from the CLS token to patch tokens.

\subsection{Few-shot Learning}

Few-shot learning aims to learn new concepts from very few labelled examples, often only one or two per class~\cite{DBLP:journals/pr/LiYMX23}. This is an important and challenging problem, as humans can learn new concepts from just a few examples whereas most machine learning models require large amounts of training data. Self-supervised techniques have gained attention in this field to facilitate performance. Transfer learning is very effective for few-shot learning~\cite{DBLP:conf/cvpr/Hu0SKH22,gidaris2020learning}, involving pretraining models on large unlabelled datasets then finetuning on the target few-shot learning dataset. 
\citeauthor{gidaris2020learning} propose an approach for improving few-shot learning through self-supervision, using SSP as an auxiliary task to learn richer and more transferable visual representations. 
To get rid of the finetuning process, researchers consider prerequisite information which is relative to the objects, such as attributes or characteristics. \citeauthor{bucher2019zero} utilise semantic word embeddings to generate unseen class embedding to obtain target segmentation. Recently, \citeauthor{kirillov2023segment} incorporate multiple prompts, including a rough box, free-form text, a set of points, or a mask, to allow zero-shot segmentation.
However, the above methods introduce external knowledge from multi-modalities or supervision to form task-relative representations. The generalization and adaptability seen in these methods might stem from the additional knowledge or the finetuning process of meta-learners. 
While these methods offer valuable insights, we propose a framework that seeks to reveal the intrinsic properties of SSP models without relying on external knowledge or meta-learners.

\section{Method}

\subsection{Problem Definition}

\begin{figure}[h]
    \centering
    \includegraphics[width=.7\linewidth]{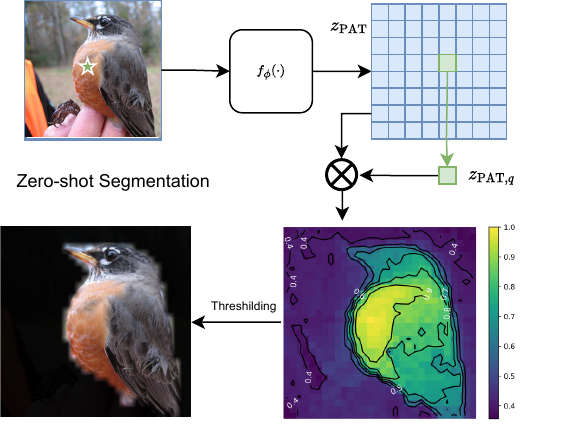}
    \caption{Overview of the proposed framework for one-shot classification and zero-shot segmentation, including feature extraction, calculating similarity, and thresholding. $\otimes$ denotes cosine similarity.
    }\label{fig:framework_a}
\end{figure}

We introduce a framework to tackle segmentation or classification based on similarity thresholding. The overview of our framework is illustrated in Figure~\ref{fig:framework_a}. 
We employ SSP ViT as an embedding function $f_\phi(\cdot)$ to extract local and global features simultaneously. ViTs typically divide a given input image $ \vx \in \mathbb{R}^{224 \times 224}$ into smaller, fixed-size, non-overlapping patches, using non-overlapping convolutions, and add position embeddings to obtain \textit{patch tokens}. 
These patch tokens, along with an extra \textit{CLS token}, are processed through transformer blocks, consisting of multiple layers of self-attention mechanisms and feed-forward networks. This process results in local features for visual patches $\vz_{\mathtt{PAT}} \in \mathbb{R}^{P \times C}$ and global features for the whole image $\vz_{\mathtt{CLS}} \in \mathbb{R}^{1 \times C}$, where $P$ and $C$ represent the number of patches and feature dimensions, respectively. 
To perform zero-shot segmentation, we measure the distance between two features $\sigma(\cdot,\cdot)$ employing cosine similarity. Then, we apply a thresholding method to perform zero-shot segmentation as discussed in the following section.

\noindent\textbf{Zero-shot Segmentation from Prompts.}\quad{}
A selected patch token $\sigma(\vz_{\mathtt{PAT},q})$ in the target object area is a prompt for segmenting the target. 
To generate higher-resolution segmentation, we re-scale the input images to 480x480, thus extracting 30x30 patch tokens for each image, given that the patch size is equal to 16x16 pixels. 
Then, we extract the patch tokens to calculate the similarity between the selected token and the rest of the tokens in the image $\sigma(\vz_{\mathtt{PAT},q},\vz_{\mathtt{PAT}})$. Finally, we get the segmentation of the given class where their similarities are higher than a certain threshold.
We segment the object region for a target object where the distance to the query patch is within $T$ for a given patch of the object.
The threshold $T$ is a hyper-parameter selected empirically.
Our setting requires no support samples with labels, following the interactive segmentation~\cite{sofiiuk2022reviving}.

This evaluation process enabled us to thoroughly examine and measure how different methods compare in terms of their inter-object and intra-object similarities. Our investigation revealed that basic MIM techniques face difficulties when used for zero-shot prompting, due to their high inter-object similarity and the self-distillation on local features can significantly enlarge the distinction between inter-object and intra-object similarities. 
Based on these findings, we propose a simple but effective SSL method, outlined in the following section.

\begin{figure}[h]
    \centering
    \includegraphics[width=.8\linewidth]{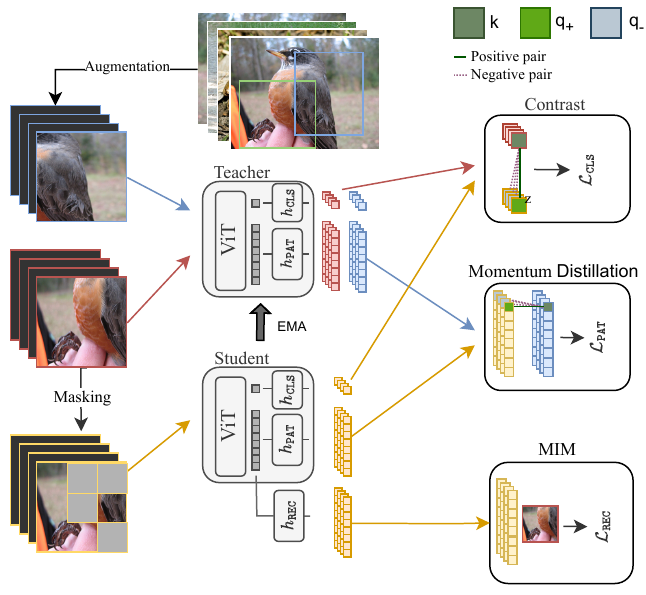}
    \caption{ The proposed MMC, performing self-supervised representation learning. The teacher takes unmasked views and the student takes masked views. MMC maximizes global agreement for the CLS token across views $\mathcal{L}_{\mathtt{CLS}}$, and local agreement for the patch tokens between masked and unmasked views $\mathcal{L}_{\mathtt{PAT}}$. In addition, MMC predicts the masked pixels $\mathcal{L}_{\mathtt{REC}}$. The masked versions are fed forward to the student network, and the complete views are fed forward to the teacher network. The parameters of the teacher network are updated from the student network using exponential moving average (EMA). The tokens from masked versions are colored gray.}\label{fig:framework_b}
\end{figure}

\subsection{Self-supervised Training}

The key observations behind the simple SSL framework we proposed are that MIM is helpful in increasing the intra-object similarity and unhelpful for inter-object discrimination, while momentum distillation via CL on local patches enhance local inter-object discrimination and effectively distill global semantic into local representations.  
Building upon these observations, we propose a simple self-supervised pretraining framework, termed Masked Momentum Contrast (MMC). 
As illustrated in Figure~\ref{fig:framework_b}, proposed framework consists of a teacher and a student as applied in~\cite{grill2020bootstrap,chen_empirical_2021,caron_emerging_2021}. Two encoders, $f_{s}$ (student encoder parameterised by $\phi_s$) and $f_t$ (teacher encoder parameterised by $\phi_t$), are used to map the input to a latent space. The student takes a masked view $\vx^m$, and the teacher takes an unmasked view $\vx$. We introduce three objectives to accomplish our goal. 

\noindent{}\textbf{Masked Image Modeling.}\quad{}
We implement a common pretext task that reconstructs pixels for masked regions~\cite{atito_gmml_2022,he_masked_2022}. 
A mean absolute loss is employed to predict the masked pixels:
 \begin{equation}
    \mathcal{L}_{\mathtt{REC}} = \frac{1}{ P} \sum_{p=1}^P \mathbf{M}_{p} \cdot \left\| \vx_p - h_\mathtt{REC}(f({\vx_{p}}^{\mathtt{m}})) \right\|,
\end{equation}
where $p$ denotes the index of pixels, $\mathbf{m}$ denotes the mask map for the masked image. 
Here MIM with pixel reconstruction is insufficient to build a high-level representation for visual patches, which has a poor performance under the linear protocol.

\noindent{}\textbf{Momentum Distillation.}\quad{}
We introduce momentum based self-distillation to transfer semantics from global features to local features, inspired by iBOT~\cite{zhou_ibot_2022}. 
The patch tokens from masked views will match the ones from unmasked views. The formula is:
\begin{equation}
\mathcal{L}_{\mathtt{PAT}} = - \frac{1}{P} \sum_{p=1}^P \log \frac{e^{q_p^{\mathtt{m}} \cdot k_{p+} / \tau}}
{e^{q_p^{\mathtt{m}} \cdot k_{p+} / \tau} + \sum_{k_{p-}} e^{q_p^{\mathtt{m}} \cdot k_{p-} / \tau}},
\end{equation}
where $q_p^{\mathtt{m}}=h_{\mathtt{PAT}}(\vz^{\mathtt{m}})$, $k_p=h_{\mathtt{PAT}}(\vz)$ are representations for the visual patch at $p$ from the masked view and the unmasked view respectively. The positive sample is the same patch from the unmasked view, and the negative samples are patches from other images in the unmasked view. 
This objective is the key component to encourage discriminative representation for segmenting objects.

\noindent{}\textbf{Global Contrast.}\quad{} 
Similar to MSN~\cite{assran2022masked_MSN}, we generate masked and unmasked views to promote invariance on the global features.
$k=h_{\mathtt{CLS}}(f_t(\vx))$. 
The teacher, taking unmasked images, supervises the student to match the masked images through InfoNCE~\cite{he_momentum_2020}:
\begin{equation}
    \mathcal{L}_\mathtt{CLS} =-\log 
    \frac
    {\exp \left(q^\mathtt{m} \cdot k_{+} / \tau\right)}
    {\exp \left(q^\mathtt{m} \cdot k_{+} / \tau\right) +
        \sum_{k_{-}} 
            \exp \left(q^\mathtt{m} \cdot k_{-} / \tau\right)}
\end{equation}
where, $\tau>0$ is a temperature parameter that controls the sharpness of the output distribution, $k^\mathtt{m}=h_{\mathtt{CLS}}(f_t(\vx^{\mathtt{m}})), q^\mathtt{m}=h_{\mathtt{CLS}}(f_s(\vx^{\mathtt{m}}))$ represents the embedding from the masked view of the input image. 
Although global contrast is proven an effective technique to learn the concept of an image, such property cannot emerge on local features naturally.

\noindent{}\textbf{Overall.}\quad{} 
$h_\mathtt{REC}(\cdot), h_\mathtt{PAT}(\cdot)$, and $h_\mathtt{CLS}(\cdot)$ are independent Multi-layer Perceptron (MLP) heads to project a common feature space to different task spaces.
Inclusion of $\mathcal{L}_\mathtt{REC}$ contributes to the acquisition of a latent semantic structure.
The incorporation of $\mathcal{L}_\mathtt{PAT}$ alongside $\mathcal{L}_\mathtt{CLS}$ not only bolsters explicit semantic representation in the CLS token but also facilitates the seamless transfer of semantics to patch tokens.
$\mathcal{L}_{\mathtt{CLS}}$ emerges as the cornerstone, exerting the most substantial influence in extracting intrinsic high-level semantics from the CLS token.
The final form is :
\begin{equation}
    \mathcal{L} =  \mathcal{L}_{\mathtt{REC}} +  \mathcal{L}_{\mathtt{CLS}} + \mathcal{L}_{\mathtt{PAT}}.
\end{equation}
Moreover, multi-crop views have proven effective in enhancing performance~\cite{zhou_ibot_2022,caron_emerging_2021}. Consequently, we introduce multi-crop views without masking for the teacher to promote semantics on global features.

\noindent\textbf{Discussion.}\quad{}
Our model differs iBOT in using a contrastive loss instead of a distillation loss, alongside introducing reconstruction pretext to be aware of subtle colour information. 
Although SiT has a compact intra-object representation, it fails to encourage the inter-object dissimilarity.
MMC improves SiT by introducing an extra loss $\mathcal{L}_{\mathtt{PAT}}$ to reduce the inter-object similarity.

\section{Experiments}

\subsection{Implementation Details}

The backbone architecture of MMC employs the small (ViT-S) and base (ViT-B) variants of ViT. 
The heads are independent and employ 3-layer MLPs with hidden size of 4096.

For pretraining, we first generate two random views/crops from the input image with size $224 \times 224$. For each view, a series of data augmentations are applied to enhance the variability of the training data. The augmentation transformations are described in Supplement.
The student network is optimized using AdamW with a weight decay of 0.04 and a cosine learning rate ranging from $7.5e^{-4}$ to $1e^{-6}$. The teacher is updated using exponential moving average of the student weights with $\lambda$ following a cosine schedule from $0.996$ to $1$ during training as in~\cite{grill_bootstrap_2020}. 
The temperature $\tau$ for contrastive learning is set to 0.2 in all scenarios.
The batch size is set to 48.
We train ViT-S/16 on ImageNet-100 (IN100) and ViT-B/16 on ImageNet-1K (IN1K)~\cite{russakovsky2015imagenet} for 500 and 400 epochs, respectively. 
IN100 is a sub set of IN1K, consisting of 100 classes, divided by ~\citeauthor{imagenet100pytorch}.

\subsection{Comparison with State-of-the-arts}

\begin{table}[h]
\centering
    
    
\begin{tabular}{@{}lcccccc}
\toprule
\multirow{2}{*}{Method}& \multicolumn{5}{c}{Threshold} & \multirow{2}{*}{Lin.}\\
\cmidrule{2-6}
& 0.1 & 0.2 & 0.3 & 0.4 & Opt. &  \\
\midrule
RAND                 & 16.8 & 16.6 & 16.2 & 15.5 & 16.8 & \\
SiT                  & 32.9 & 35.3 & 34.7 & 31.6 & 35.3 & 34.7                  \\
DINO                 & 33.3 & 35.1 & 33.0 & 28.2 & 35.1 & 42.7                  \\
iBOT                 & 33.5 & 36.2 & 36.2 & 34.1 & 36.2 & 45.3                  \\
MC-SSL               & 31.9 & 34.2 & 35.0 & 35.1 & 35.1 & 39.7                  \\
MAE                  & 28.6 & 30.1 & 26.0 & 18.5 & 30.1 & 24.6                  \\
MSN                  & 32.7 & 34.4 & 31.6 & 25.9 & 34.4 & 42.2                    \\
MMC (ours)           & \textbf{34.6} & \textbf{37.6} & \textbf{38.4} & \textbf{37.9} & \textbf{38.4} & \textbf{46.4}                  \\
\bottomrule
\end{tabular}
\caption{Semantic segmentation (mIoU)  with thresholding from 0.1 to 0.4 and linear frozen features (Lin.) on COCO. ``Opt.'' denotes the performance on optimal threshold sweeping from 0 to 0.9. The architecture for SSL models is ViT-B.}
\label{tab:coco_seg}
\end{table}

\noindent\textbf{Zero-shot Segmentation on COCO.}\quad{}
We employed the Common Objects in Context (COCO) dataset~\cite{lin2014COCO} to evaluate zero-shot segmentation. With its extensive collection of over 200,000 images featuring more than 500,000 annotated objects distributed across 91 distinct categories, COCO presents a robust platform for our investigation. 
Through a systematic exploration, we varied the threshold parameter across the range of 0 to 0.3, increasing it by intervals of 0.1. The performance of SSL models (ViT-B) as well as SOTA self-supervised segmentation models, quantified by the mean Intersection over Union (mIoU), is documented for each threshold value in Table~\ref{tab:coco_seg}. 
Impressively, our model achieves the highest mIoU score of 38.4\% with the adoption of a solitary threshold, highlighting its superiority on extracting semantic information. 
In addition, we also report the accuracy for finetuning the entire network in Table~\ref{tab:coco_seg}.
The experimental results demonstrate that MMC achieves the best segmentation performance with a large margin on both the zero-shot setting and training a linear head only.

\begin{table}

\centering

\begin{tabular}{@{}lcccc@{}}
\toprule
                                    & ARCH & $\mathcal{J\&F}_m$  & $\mathcal{J}_m$ & $\mathcal{F}_m$ \\
\midrule
\multicolumn{4}{l}{\textit{Trained on IN100}} \\
Supervised                             & ViT-S/16    & 37.1     & 35.8  & 38.5  \\
MAE                                   & ViT-S/16    & 44.3     & 43.6  & 45.1  \\
MSN                                   & ViT-S/16    & 48.9     & 46.9  & 51.0  \\
DINO                                  & ViT-S/16    & 57.7     & 56.2  & 59.1  \\
iBOT                                  & ViT-S/16    & 59.7     & 57.9  & 61.4  \\
MMC (ours)                             & ViT-S/16    & \textbf{60.3}     & \textbf{58.6}  & \textbf{62.0}  \\
\midrule
\multicolumn{4}{l}{\textit{Trained on IN1K}} \\
DINO                                  & ViT-B/16    & 62.3     & 60.7  & 63.9  \\
MC-SSL                                & ViT-B/16    & 60.4     & 58.6  & 62.2  \\
MAE                                   & ViT-B/16    & 47.4     & 46.1  & 48.8  \\
MSN                                   & ViT-B/16    & 61.4     & 59.6  & 63.3  \\
MMC (ours)                                   & ViT-B/16    & \textbf{63.8}     & \textbf{61.8}  & \textbf{65.6} \\
\bottomrule
\end{tabular}
\caption{Video object segmentation by nearest neighbor retrieval from pre-trained features on DAVIS (480p). We report the mean region similarity (J-Mean) and mean contour-based accuracy (F-Mean). w/o denotes without multi-crops.}   \label{tab:video_seg}
\vspace{-1em}
\end{table}
\noindent{}\textbf{Video Segmentation on DAVIS-2017.}\quad{}
Learned features from ViTs have demonstrated their strong potential for downstream tasks through nearest retrieval~\cite{zhou_ibot_2022,caron_emerging_2021}. In accordance with the evaluation protocol established by ~\cite{jabri2020space}, we employ frozen pretrained features to compute cosine similarity among features. This approach enables label identification through nearest neighbor analysis between consecutive frames, subsequently propagating masks.
To validate the proficiency of our patch features, we subject them to evaluation using the DAVIS-2017 video instance segmentation benchmark~\cite{pont2017davis}. Our model's performance, as highlighted in Table~\ref{tab:video_seg}, consistently outperforms other models, both with ViT-S and ViT-B architectures.
Notably, it is observed that MMC achieves a substantial margin of improvement when trained on a larger dataset. This outcome underscores the potency of our model when applied to large-scale datasets.

\begin{table}[t]
\centering

\setlength{\tabcolsep}{5pt}
\begin{tabular}{lcccc}
\toprule
\multirow{2}{*}{Methods} &  \multirow{2}{*}{\begin{tabular}{c} Backbone \\ (\# params) \end{tabular} } & \multicolumn{3}{c}{Datasets}          \\ 
\cmidrule{3-5} 
  &    & \scriptsize{ADE20k}  & \scriptsize{Cityscapes} & \scriptsize{VOC aug}\\ 
  \midrule
Random &  \multirow{8}{*}{\breakrows{ ViT-B/16 \\ (85M) } }& 2.1 &13.7 &4.7\\
MoCo-V3 &  & 28.7 & -- & 64.8\\

SiT   & & 33.8 & 56.1 & 74.8          \\
DINO  & & 32.6 &  59.1 & 73.5          \\
iBOT  &  & 34.7 &  60.7 & 75.6          \\
MAE  & & 24.3 & 44.2 & 58.4\\
MC-SSL  & & 33.2 & 59.7 & 75.9\\
MSN   &  & 32.9 & 55.1 & 72.2          \\
MMC (ours) & & \textbf{37.0} & \textbf{62.6}      & \textbf{77.7} \\ 
\bottomrule
\end{tabular}
\caption{Semantic segmentation (mIoU) with linear frozen features on cityscapes, VOC aug, and ADE20K. All models are pretrained on IN1K.``Random'' denotes models with randomly initialized parameters.}\label{tab:seg}
\vspace{-1em}
\end{table}

\noindent{}\textbf{Linear Frozen Segmentation.}\quad{}
Following the prescribed configurations outlined by \cite{bardes2022vicregl} for consistency, we direct the frozen patch features extracted from layers 6, 8, 10, and 12 through a trainable linear layer to adapt them to the requirements of semantic segmentation tasks. Our training process encompasses 40,000 iterations for each individual model, carried out on the ADE20K~\cite{zhou2019semanticADE}, CityScapes~\cite{cordts2016cityscapes}, and PASCAL VOC 2012~\cite{everingham2010pascal}.
Table \ref{tab:seg} presents the outcomes, where it becomes evident that our model excels across all three datasets.
Notably, our approach, MMC, stands out with remarkable mIoU scores of 37.0 on ADE20K, 62.6 on Cityscapes, and an impressive 77.7 on VOC augmentation. These results solidify MMC's prowess in semantic segmentation tasks.

\begin{figure}[thb]
    \centering
    \includegraphics[width=.73\linewidth]{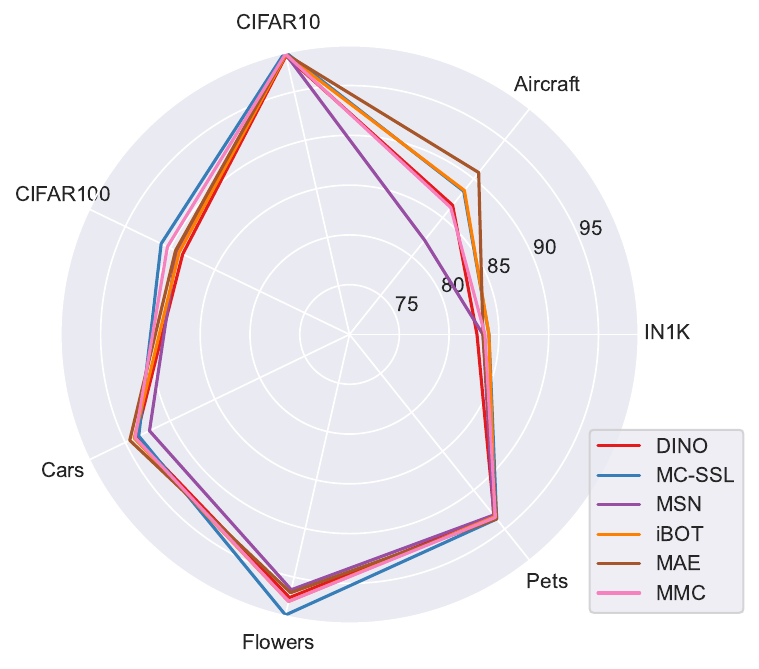}
    \caption{Transfer learning by finetuning pretrained models on different datasets. We report top-1 accuracy. No SSL model excels others on all datasets.}
    \label{fig:transfer}
\end{figure}

\noindent\textbf{Transfer Learning.}\quad
We evaluate the the quality of the features pretrained with SSL models on different downstream tasks. We use the same finetune recipe for all datasets, see details in Appendix. We compare with features from the ViT-B/16 pretrained on IN1K in Figure~\ref{fig:transfer}. We observe that finetuning performance is inconsistent to the linear protocol, which is not suitable for evaluating the representations. Moreover, MMC excels others on Flowers, maintaining competitive performance on other datasets.

\section{Experimental Analysis}

\begin{table}[tbh]

\centering

\begin{tabular}{cccccc}
\toprule
$\mathcal{L}_{\mathtt{REC}}$ & $\mathcal{L}_{\mathtt{CLS}}$ & $\mathcal{L}_{\mathtt{PAT}}$ &  KNN$_{\mathtt{CLS}}$   & KNN$_{\mathtt{PAT}}$    & Fin$_{\mathtt{CLS,PAT}}$   \\
\midrule
\cmark & \xmark & \xmark & 17.18    & 14.70              & 89.94                   \\
\xmark & \cmark & \xmark & 75.02    & 63.88             & \textbf{90.62}           \\
\xmark & \xmark & \cmark & 39.92    & 40.82             & 87.84                     \\
\midrule
\xmark & \cmark & \cmark & 75.56    & 75.54             & \textbf{89.72}        \\ 
\cmark & \xmark & \cmark & 33.54    & 33.74             & 89.40                    \\
\cmark & \cmark & \xmark & 72.54    & 53.68             & 88.24                   \\
\midrule

\cmark & \cmark & \cmark & \textbf{77.54}    & \textbf{76.96}             & \textbf{91.02}                  \\
\bottomrule
\end{tabular}
\caption{Classification performance (accuracy) of ViT-S on IN100 for different combinations of used objects. KNN$_{\mathtt{CLS}}$ utilises the CLS token, KNN$_{\mathtt{PAT}}$ utilises the mean of patch tokens, Fin$_{\mathtt{CLS,PAT}}$ concatenates the CLS and the mean of patch tokens.}\label{tab:ablation}
\vspace{-1em}


\end{table}

\noindent\textbf{Ablation of Components.}\quad{}
In this part, we study the various components introduced within our model to unravel their individual functions and contributions. Our model was trained utilizing ViT-S/16 architecture on the IN100 dataset over 200 epochs. We evaluate performance of classification through both K-nearest neighbor (KNN) and finetuning on IN100. 
The presented Table~\ref{tab:ablation} embodies the outcomes of our ablation study, underscoring the impacts of the three distinct objectives, which derives insights:
1) The removal of $\mathcal{L}_\mathtt{CLS}$ substantially impacts the performance, resulting in a noteworthy 44\% decline in KNN$_{\mathrm{CLS}}$ accuracy. This underscores the pivotal role of $\mathcal{L}_\mathtt{CLS}$ in extracting high-level semantics from the CLS token.
2) The introduction of $\mathcal{L}_\mathtt{PAT}$ alongside $\mathcal{L}_\mathtt{CLS}$ yields a notable improvement of +11\% in KNN$_{\mathrm{PAT}}$ accuracy. This indicates that $\mathcal{L}_\mathtt{PAT}$ facilitates the transfer of learned semantics from the CLS token to patch tokens, in line with the findings in ~\citeauthor{zhou_ibot_2022}.
3) While $\mathcal{L}_\mathtt{REC}$ exhibits lower accuracy for both CLS and patch tokens in the context of KNN, it significantly enhances finetune performance. This suggests that $\mathcal{L}_\mathtt{REC}$ serves to foster an underlying semantic structure which performs poorly without finetuning. 
In summation, our model inherits the strengths of SSL techniques. It accomplishes this by engendering underlying semantic structures via $\mathcal{L}_\mathtt{REC}$, facilitating explicit semantic representation through $\mathcal{L}_\mathtt{CLS}$, and transferring semantics to patch tokens via $\mathcal{L}_\mathtt{PAT}$.

\begin{figure}[b]
    \centering

    \centering
    \includegraphics[width=.9\linewidth]{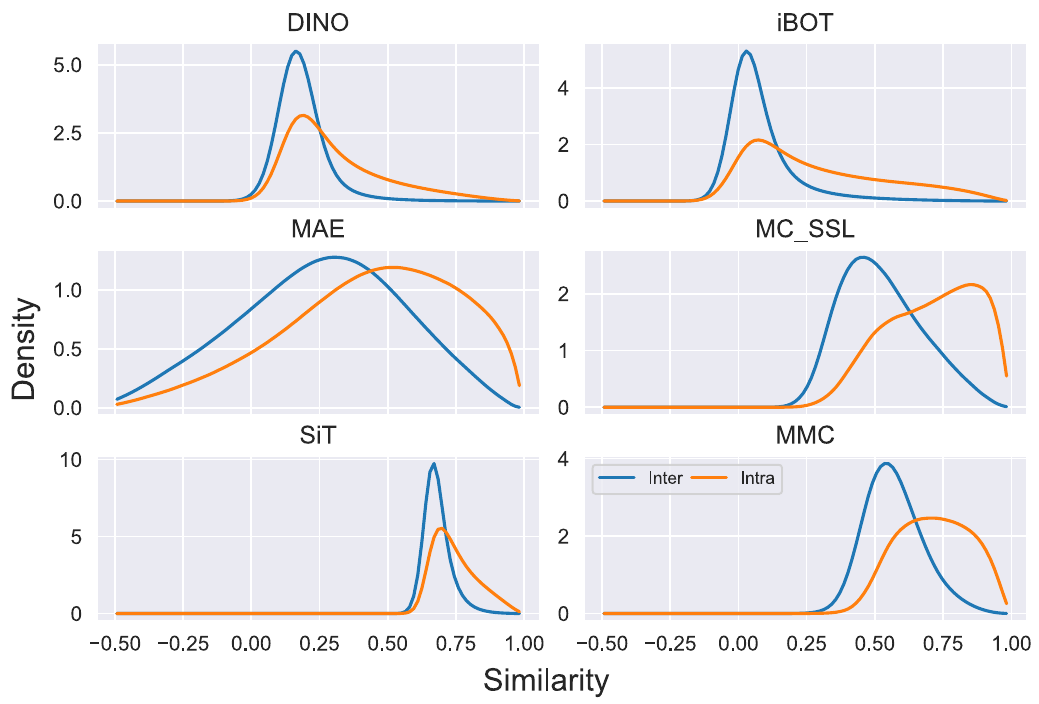}
    \caption{Intra-object and inter-object similarity distribution for SSL models.}
    \label{fig:sim_dist}
    \vspace{-0.5em}

\end{figure}

\noindent{}\textbf{Similarity Distribution.}\quad{}
\label{sec:similarity}
We define the \textit{intra-object patches} as patch pairs belonging to the same category, while \textit{inter-object patches} are pairs belonging to different categories. Ideally, their similarity distributions should be separable and have a low overlap area, ensuring distinct and identifiable features for segmentation. We show the intra- and inter- object similarity distributions in Figure~\ref{fig:sim_dist} and the statistics of these distributions in Table~\ref{tab:overlap}. 
We can see that 1) View-invariance based methods (DINO, iBOT) have lower inter-object similarity. 
2) Momentum based self-distillation is effective to encourage the representation discrimination (iBOT, MC-SSL, MMC).
3) The reconstruction pretext (SiT, MAE, MC-SSL, MMC) brings a negative impact on representation learning, leading to high inter-object similarities.
Among them, SiT excels other in encouraging compactness of intra-object representations yet has high inter-object similarity.
By introducing $\mathcal{L}_\mathtt{PAT}$, our model relieves this impact and achieves the best balance such that MMC has the lowest overlap area of 0.55. 

\begin{table}[]
\scalebox{0.94}{
    \begin{tabular}{ccccccc}
    \toprule
     &  DINO & iBOT          & SiT  & MAE  & MC-SSL & MMC          \\
           \midrule
 O     & 0.66 & 0.56          & 0.63 & 0.76 & 0.58   & \textbf{0.55} \\
 Intra & 0.2  & 0.19          & \textbf{0.5}  & 0.29 & 0.47   & 0.48 \\
 Inter & 0.12 & \textbf{0.05} & 0.46 & 0.17 & 0.35   & 0.38     \\
     \bottomrule
    \end{tabular}
}
    \caption{Statistics of similarity distribution. ``O'' denotes the overlap area of intra- and inter- object distributions, which should be small. ``Intra, Inter'' denote the mean of intra- and inter- object distributions respectively.}
    \label{tab:overlap}
\end{table}

\begin{figure}[!hbt]
    \centering
    \includegraphics[width=0.95\linewidth]{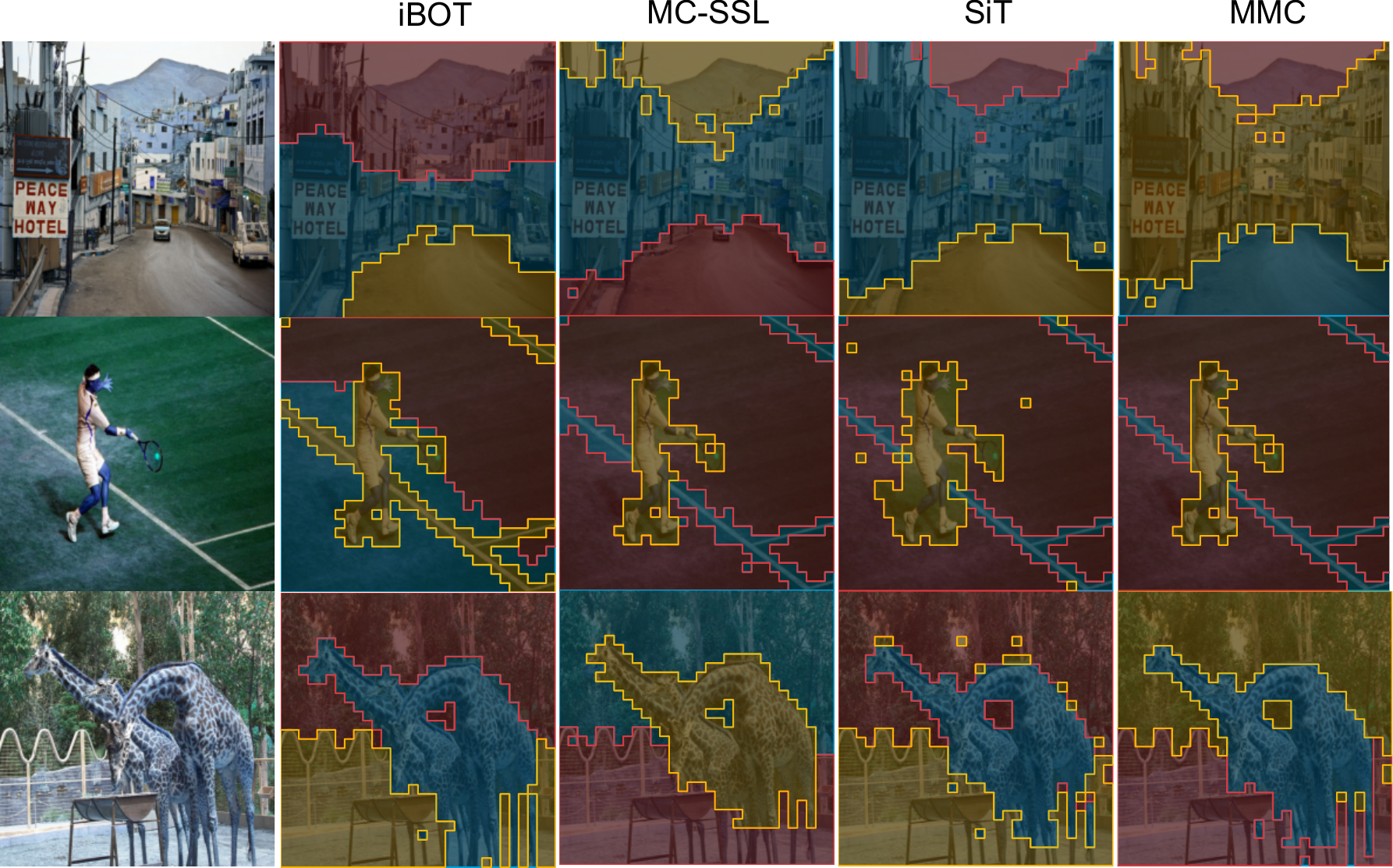}
    \caption{Cluster visualization through k-means (3 clusters).}
    \label{fig:clusters}
\end{figure}

\noindent\textbf{Segmentation Samples.}\quad{}
We also show the visualization of feature clusters for patch tokens through k-means (3) in Figure~\ref{fig:clusters}. For the first example, the mountain in the background has a close color to the buildings beside the street, where iBOT cannot distinguish them well. For the second example, MMC separates the sportsman, lines, and ground well. Although iBOT has shown impressive performance in linear evaluation, its features are still in a low-level intra-object similarity. From above examples, we can see that iBOT is not good at dealing with color information which probably is dropped for distillation only. 

\begin{figure}[!hbt]
    \centering
    \includegraphics[width=.78\linewidth]{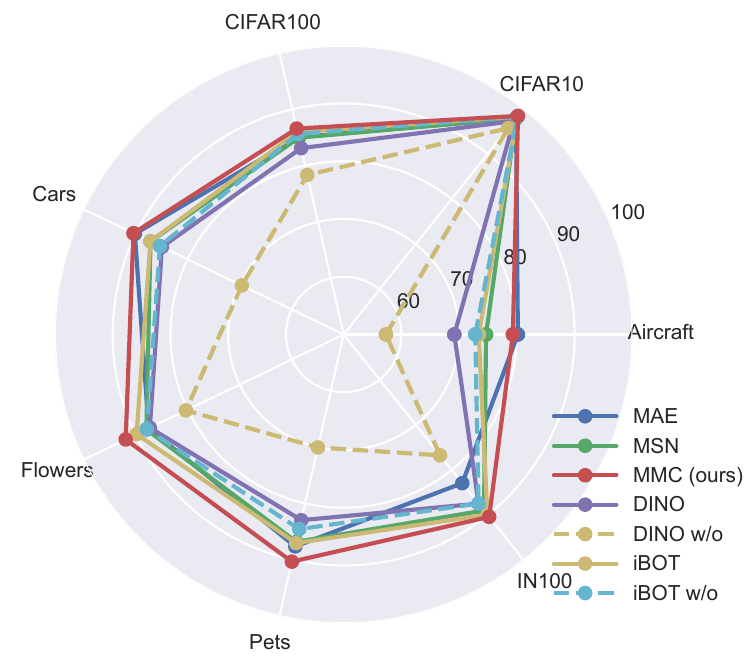}
    \caption{Finetuning performance of ViT/S pretrained on IN100. w/o denotes the model without multi-crops.}
    \label{fig:vit-s_IN100}
\end{figure}

\noindent{}\textbf{Data Efficiency.}\quad{}
We pretrain SSL methods with ViT-S/16 on the IN100 dataset for 400 epochs except MAE for 800 epochs.
After pretraining, the models are finetuned on seven different datasets, including Flowers, Pets, Cars, CIFAR10, CIFAR100, and IN100. The results, as shown in Figure~\ref{fig:vit-s_IN100}, indicate that our method consistently outperformed the other methods across all datasets, including the pretrained models with local crops, denoted as (w/o). Notably, the performance of our model consistently covers other models. These results highlight the capability of our method to efficiently learn high-quality representations, even when the pretraining is conducted on small datasets.

\section{Conclusion}

This work develops a novel evaluation protocol for SSL models in terms of addressing zero-shot segmentation by prompting one single patch belonging to the target object through a simple similarity thresholding algorithm.
We investigate the reason for the poor linear separability of learned representations by MIM methods, which is mainly raised by their high inter-object similarity.
Therefore, we propose MMC to apply the momentum distillation to diminish this issue, which significantly reduces the overlap area of inter- and intra- object similarity distribution.
We view MMC as an opportunity for future research to delve deeper into this issue and devise novel strategies to enhance inter-object distinction in SSL.
Limited by the ViTs architecture, our algorithm can not produce high-resolution segmentation masks, can be probably solved with the use of hierarchical ViTs with a smaller patch size.

\noindent\textbf{Broader Impact. }
Our analysis of similarity distributions provides insights into improving SSL models for learning object-discriminative features. We urge the community to be aware and identify the low discriminatory property of the reconstruction pretext for representation learning.

\clearpage

\bibliography{references}

\begin{thebibliography}{55}
\providecommand{\natexlab}[1]{#1}

\bibitem[{Assran et~al.(2022)Assran, Caron, Misra, Bojanowski, Bordes, Vincent,
  Joulin, Rabbat, and Ballas}]{assran2022masked_MSN}
Assran, M.; Caron, M.; Misra, I.; Bojanowski, P.; Bordes, F.; Vincent, P.;
  Joulin, A.; Rabbat, M.; and Ballas, N. 2022.
\newblock Masked siamese networks for label-efficient learning.
\newblock In \emph{Computer Vision--ECCV 2022: 17th European Conference, Tel
  Aviv, Israel, October 23--27, 2022, Proceedings, Part XXXI}, 456--473.
  Springer.

\bibitem[{Atito et~al.(2021)Atito, Awais, Farooq, Feng, and
  Kittler}]{atito_mc-ssl00_2021}
Atito, S.; Awais, M.; Farooq, A.; Feng, Z.; and Kittler, J. 2021.
\newblock {MC}-{SSL}0.0: Towards Multi-Concept Self-Supervised Learning.
\newblock \emph{ArXiv preprint}, abs/2111.15340.

\bibitem[{Atito, Awais, and Kittler(2021)}]{atito_sit_2021}
Atito, S.; Awais, M.; and Kittler, J. 2021.
\newblock {SiT}: Self-supervised {vIsion} Transformer.
\newblock \emph{ArXiv preprint}, abs/2104.03602.

\bibitem[{Atito, Awais, and Kittler(2022)}]{atito_gmml_2022}
Atito, S.; Awais, M.; and Kittler, J. 2022.
\newblock {GMML} is All you Need.
\newblock \emph{ArXiv preprint}, abs/2205.14986.

\bibitem[{Bao et~al.(2021)Bao, Dong, Piao, and Wei}]{bao_beit_2022}
Bao, H.; Dong, L.; Piao, S.; and Wei, F. 2021.
\newblock {BEiT}: {BERT} Pre-Training of Image Transformers.
\newblock \emph{ArXiv preprint}, abs/2106.08254.

\bibitem[{Bardes, Ponce, and LeCun(2021)}]{bardes2021vicreg}
Bardes, A.; Ponce, J.; and LeCun, Y. 2021.
\newblock Vicreg: Variance-invariance-covariance regularization for
  self-supervised learning.
\newblock \emph{arXiv preprint arXiv:2105.04906}.

\bibitem[{Bardes, Ponce, and LeCun(2022)}]{bardes2022vicregl}
Bardes, A.; Ponce, J.; and LeCun, Y. 2022.
\newblock VICRegL: Self-Supervised Learning of Local Visual Features.
\newblock arXiv:2210.01571.

\bibitem[{Brown et~al.(2020)Brown, Mann, Ryder, Subbiah, Kaplan, Dhariwal,
  Neelakantan, Shyam, Sastry, Askell et~al.}]{brown2020language}
Brown, T.; Mann, B.; Ryder, N.; Subbiah, M.; Kaplan, J.~D.; Dhariwal, P.;
  Neelakantan, A.; Shyam, P.; Sastry, G.; Askell, A.; et~al. 2020.
\newblock Language models are few-shot learners.
\newblock \emph{Advances in neural information processing systems}, 33:
  1877--1901.

\bibitem[{Bucher et~al.(2019)Bucher, Vu, Cord, and P{\'e}rez}]{bucher2019zero}
Bucher, M.; Vu, T.-H.; Cord, M.; and P{\'e}rez, P. 2019.
\newblock Zero-shot semantic segmentation.
\newblock \emph{Advances in Neural Information Processing Systems}, 32.

\bibitem[{Caron et~al.(2021)Caron, Touvron, Misra, Jégou, Mairal, Bojanowski,
  and Joulin}]{caron_emerging_2021}
Caron, M.; Touvron, H.; Misra, I.; Jégou, H.; Mairal, J.; Bojanowski, P.; and
  Joulin, A. 2021.
\newblock Emerging Properties in Self-Supervised Vision Transformers.
\newblock \emph{ArXiv preprint}, abs/2104.14294.

\bibitem[{Chen et~al.(2020{\natexlab{a}})Chen, Radford, Child, Wu, Jun, Luan,
  and Sutskever}]{chen2020generative}
Chen, M.; Radford, A.; Child, R.; Wu, J.; Jun, H.; Luan, D.; and Sutskever, I.
  2020{\natexlab{a}}.
\newblock Generative pretraining from pixels.
\newblock In \emph{International conference on machine learning}, 1691--1703.
  PMLR.

\bibitem[{Chen et~al.(2020{\natexlab{b}})Chen, Kornblith, Norouzi, and
  Hinton}]{chen_simple_2020}
Chen, T.; Kornblith, S.; Norouzi, M.; and Hinton, G.~E. 2020{\natexlab{b}}.
\newblock A Simple Framework for Contrastive Learning of Visual
  Representations.
\newblock In \emph{Proceedings of the 37th International Conference on Machine
  Learning, {ICML} 2020, 13-18 July 2020, Virtual Event}, volume 119 of
  \emph{Proceedings of Machine Learning Research}, 1597--1607. {PMLR}.

\bibitem[{Chen et~al.(2020{\natexlab{c}})Chen, Kornblith, Swersky, Norouzi, and
  Hinton}]{DBLP:conf/nips/ChenKSNH20}
Chen, T.; Kornblith, S.; Swersky, K.; Norouzi, M.; and Hinton, G.~E.
  2020{\natexlab{c}}.
\newblock Big Self-Supervised Models are Strong Semi-Supervised Learners.
\newblock In Larochelle, H.; Ranzato, M.; Hadsell, R.; Balcan, M.; and Lin, H.,
  eds., \emph{Advances in Neural Information Processing Systems 33: Annual
  Conference on Neural Information Processing Systems 2020, NeurIPS 2020,
  December 6-12, 2020, virtual}.

\bibitem[{Chen, Luo, and Li(2021)}]{chen2021intriguing}
Chen, T.; Luo, C.; and Li, L. 2021.
\newblock Intriguing properties of contrastive losses.
\newblock \emph{Advances in Neural Information Processing Systems}, 34:
  11834--11845.

\bibitem[{Chen et~al.(2022)Chen, Ding, Wang, Xin, Mo, Wang, Han, Luo, Zeng, and
  Wang}]{chen_context_2022}
Chen, X.; Ding, M.; Wang, X.; Xin, Y.; Mo, S.; Wang, Y.; Han, S.; Luo, P.;
  Zeng, G.; and Wang, J. 2022.
\newblock Context Autoencoder for Self-Supervised Representation Learning.
\newblock \emph{arXiv preprint arXiv:2202.03026}.

\bibitem[{Chen, Xie, and He(2021)}]{chen_empirical_2021}
Chen, X.; Xie, S.; and He, K. 2021.
\newblock An Empirical Study of Training Self-Supervised Vision Transformers.
\newblock \emph{ArXiv preprint}, abs/2104.02057.

\bibitem[{Chun-Hsiao~Yeh(2022)}]{imagenet100pytorch}
Chun-Hsiao~Yeh, Y.~C. 2022.
\newblock {IN100pytorch}: PyTorch Implementation: Training ResNets on
  ImageNet-100.
\newblock \url{https://github.com/danielchyeh/ImageNet-100-Pytorch}.

\bibitem[{Cordts et~al.(2016)Cordts, Omran, Ramos, Rehfeld, Enzweiler,
  Benenson, Franke, Roth, and Schiele}]{cordts2016cityscapes}
Cordts, M.; Omran, M.; Ramos, S.; Rehfeld, T.; Enzweiler, M.; Benenson, R.;
  Franke, U.; Roth, S.; and Schiele, B. 2016.
\newblock The cityscapes dataset for semantic urban scene understanding.
\newblock In \emph{Proceedings of the IEEE conference on computer vision and
  pattern recognition}, 3213--3223.

\bibitem[{Devlin et~al.(2018)Devlin, Chang, Lee, and
  Toutanova}]{devlin2018bert}
Devlin, J.; Chang, M.-W.; Lee, K.; and Toutanova, K. 2018.
\newblock Bert: Pre-training of deep bidirectional transformers for language
  understanding.
\newblock \emph{arXiv preprint arXiv:1810.04805}.

\bibitem[{Dosovitskiy et~al.(2020)Dosovitskiy, Beyer, Kolesnikov, Weissenborn,
  Zhai, Unterthiner, Dehghani, Minderer, Heigold, Gelly
  et~al.}]{dosovitskiy2020vit}
Dosovitskiy, A.; Beyer, L.; Kolesnikov, A.; Weissenborn, D.; Zhai, X.;
  Unterthiner, T.; Dehghani, M.; Minderer, M.; Heigold, G.; Gelly, S.; et~al.
  2020.
\newblock An image is worth 16x16 words: Transformers for image recognition at
  scale.
\newblock \emph{arXiv preprint arXiv:2010.11929}.

\bibitem[{El{-}Nouby et~al.(2021)El{-}Nouby, Neverova, Laptev, and
  J{\'{e}}gou}]{DBLP:journals/corr/IRT}
El{-}Nouby, A.; Neverova, N.; Laptev, I.; and J{\'{e}}gou, H. 2021.
\newblock Training Vision Transformers for Image Retrieval.
\newblock \emph{CoRR}, abs/2102.05644.

\bibitem[{Everingham et~al.(2010)Everingham, Van~Gool, Williams, Winn, and
  Zisserman}]{everingham2010pascal}
Everingham, M.; Van~Gool, L.; Williams, C.~K.; Winn, J.; and Zisserman, A.
  2010.
\newblock The pascal visual object classes (voc) challenge.
\newblock \emph{International journal of computer vision}, 88: 303--338.

\bibitem[{Gidaris et~al.(2020)Gidaris, Bursuc, Komodakis, P{\'e}rez, and
  Cord}]{gidaris2020learning}
Gidaris, S.; Bursuc, A.; Komodakis, N.; P{\'e}rez, P.; and Cord, M. 2020.
\newblock Learning representations by predicting bags of visual words.
\newblock In \emph{Proceedings of the IEEE/CVF Conference on Computer Vision
  and Pattern Recognition}, 6928--6938.

\bibitem[{Grill et~al.(2020{\natexlab{a}})Grill, Strub, Altch{\'{e}}, Tallec,
  Richemond, Buchatskaya, Doersch, Pires, Guo, Azar, Piot, Kavukcuoglu, Munos,
  and Valko}]{grill_bootstrap_2020}
Grill, J.; Strub, F.; Altch{\'{e}}, F.; Tallec, C.; Richemond, P.~H.;
  Buchatskaya, E.; Doersch, C.; Pires, B.~{\'{A}}.; Guo, Z.; Azar, M.~G.; Piot,
  B.; Kavukcuoglu, K.; Munos, R.; and Valko, M. 2020{\natexlab{a}}.
\newblock Bootstrap Your Own Latent - {A} New Approach to Self-Supervised
  Learning.
\newblock In Larochelle, H.; Ranzato, M.; Hadsell, R.; Balcan, M.; and Lin, H.,
  eds., \emph{Advances in Neural Information Processing Systems 33: Annual
  Conference on Neural Information Processing Systems 2020, NeurIPS 2020,
  December 6-12, 2020, virtual}.

\bibitem[{Grill et~al.(2020{\natexlab{b}})Grill, Strub, Altch{\'e}, Tallec,
  Richemond, Buchatskaya, Doersch, Avila~Pires, Guo, Gheshlaghi~Azar
  et~al.}]{grill2020bootstrap}
Grill, J.-B.; Strub, F.; Altch{\'e}, F.; Tallec, C.; Richemond, P.;
  Buchatskaya, E.; Doersch, C.; Avila~Pires, B.; Guo, Z.; Gheshlaghi~Azar, M.;
  et~al. 2020{\natexlab{b}}.
\newblock Bootstrap your own latent-a new approach to self-supervised learning.
\newblock \emph{Advances in neural information processing systems}, 33:
  21271--21284.

\bibitem[{He et~al.(2022)He, Chen, Xie, Li, Dollar, and
  Girshick}]{he_masked_2022}
He, K.; Chen, X.; Xie, S.; Li, Y.; Dollar, P.; and Girshick, R. 2022.
\newblock Masked Autoencoders Are Scalable Vision Learners.
\newblock In \emph{2022 {IEEE}/{CVF} Conference on Computer Vision and Pattern
  Recognition ({CVPR})}, 15979--15988. {IEEE}.
\newblock ISBN 978-1-66546-946-3.

\bibitem[{He et~al.(2020)He, Fan, Wu, Xie, and Girshick}]{he_momentum_2020}
He, K.; Fan, H.; Wu, Y.; Xie, S.; and Girshick, R.~B. 2020.
\newblock Momentum Contrast for Unsupervised Visual Representation Learning.
\newblock In \emph{2020 {IEEE/CVF} Conference on Computer Vision and Pattern
  Recognition, {CVPR} 2020, Seattle, WA, USA, June 13-19, 2020}, 9726--9735.
  {IEEE}.

\bibitem[{Hu et~al.(2022)Hu, Li, St{\"{u}}hmer, Kim, and
  Hospedales}]{DBLP:conf/cvpr/Hu0SKH22}
Hu, S.~X.; Li, D.; St{\"{u}}hmer, J.; Kim, M.; and Hospedales, T.~M. 2022.
\newblock Pushing the Limits of Simple Pipelines for Few-Shot Learning:
  External Data and Fine-Tuning Make a Difference.
\newblock In \emph{{IEEE/CVF} Conference on Computer Vision and Pattern
  Recognition, {CVPR} 2022, New Orleans, LA, USA, June 18-24, 2022},
  9058--9067. {IEEE}.

\bibitem[{Huang et~al.(2022)Huang, Jin, Lu, Hou, Cheng, Fu, Shen, and
  Feng}]{huang2022contrastive}
Huang, Z.; Jin, X.; Lu, C.; Hou, Q.; Cheng, M.-M.; Fu, D.; Shen, X.; and Feng,
  J. 2022.
\newblock Contrastive masked autoencoders are stronger vision learners.
\newblock \emph{arXiv preprint arXiv:2207.13532}.

\bibitem[{Jabri, Owens, and Efros(2020)}]{jabri2020space}
Jabri, A.; Owens, A.; and Efros, A. 2020.
\newblock Space-time correspondence as a contrastive random walk.
\newblock \emph{Advances in neural information processing systems}, 33:
  19545--19560.

\bibitem[{Kirillov et~al.(2023)Kirillov, Mintun, Ravi, Mao, Rolland, Gustafson,
  Xiao, Whitehead, Berg, Lo et~al.}]{kirillov2023segment}
Kirillov, A.; Mintun, E.; Ravi, N.; Mao, H.; Rolland, C.; Gustafson, L.; Xiao,
  T.; Whitehead, S.; Berg, A.~C.; Lo, W.-Y.; et~al. 2023.
\newblock Segment anything.
\newblock \emph{arXiv preprint arXiv:2304.02643}.

\bibitem[{Krizhevsky, Hinton et~al.(2009)}]{krizhevsky2009learning_cifar10}
Krizhevsky, A.; Hinton, G.; et~al. 2009.
\newblock Learning multiple layers of features from tiny images.

\bibitem[{Li et~al.(2023)Li, Yang, Ma, and Xue}]{DBLP:journals/pr/LiYMX23}
Li, X.; Yang, X.; Ma, Z.; and Xue, J. 2023.
\newblock Deep metric learning for few-shot image classification: {A} Review of
  recent developments.
\newblock \emph{Pattern Recognit.}, 138: 109381.

\bibitem[{Lin et~al.(2014)Lin, Maire, Belongie, Hays, Perona, Ramanan,
  Doll{\'a}r, and Zitnick}]{lin2014COCO}
Lin, T.-Y.; Maire, M.; Belongie, S.; Hays, J.; Perona, P.; Ramanan, D.;
  Doll{\'a}r, P.; and Zitnick, C.~L. 2014.
\newblock Microsoft coco: Common objects in context.
\newblock In \emph{Computer Vision--ECCV 2014: 13th European Conference,
  Zurich, Switzerland, September 6-12, 2014, Proceedings, Part V 13}, 740--755.
  Springer.

\bibitem[{Mo, Sun, and Li(2021)}]{mo2021spcl}
Mo, S.; Sun, Z.; and Li, C. 2021.
\newblock Siamese Prototypical Contrastive Learning.
\newblock In \emph{Proceedings of British Machine Vision Conference}.

\bibitem[{Mo, Sun, and Li(2022)}]{mo2022pauc}
Mo, S.; Sun, Z.; and Li, C. 2022.
\newblock Rethinking Prototypical Contrastive Learning through Alignment,
  Uniformity and Correlation.
\newblock In \emph{Proceedings of British Machine Vision Conference}.

\bibitem[{Parkhi et~al.(2012)Parkhi, Vedaldi, Zisserman, and
  Jawahar}]{parkhi2012cats}
Parkhi, O.~M.; Vedaldi, A.; Zisserman, A.; and Jawahar, C. 2012.
\newblock Cats and dogs.
\newblock In \emph{2012 IEEE conference on computer vision and pattern
  recognition}, 3498--3505. IEEE.

\bibitem[{Pont-Tuset et~al.(2017)Pont-Tuset, Perazzi, Caelles, Arbel{\'a}ez,
  Sorkine-Hornung, and Van~Gool}]{pont2017davis}
Pont-Tuset, J.; Perazzi, F.; Caelles, S.; Arbel{\'a}ez, P.; Sorkine-Hornung,
  A.; and Van~Gool, L. 2017.
\newblock The 2017 davis challenge on video object segmentation.
\newblock \emph{arXiv preprint arXiv:1704.00675}.

\bibitem[{Radenovic et~al.(2018)Radenovic, Iscen, Tolias, Avrithis, and
  Chum}]{DBLP:conf/cvpr/RadenovicITAC18}
Radenovic, F.; Iscen, A.; Tolias, G.; Avrithis, Y.; and Chum, O. 2018.
\newblock Revisiting Oxford and Paris: Large-Scale Image Retrieval
  Benchmarking.
\newblock In \emph{2018 {IEEE} Conference on Computer Vision and Pattern
  Recognition, {CVPR} 2018, Salt Lake City, UT, USA, June 18-22, 2018},
  5706--5715. Computer Vision Foundation / {IEEE} Computer Society.

\bibitem[{Ramesh et~al.(2021)Ramesh, Pavlov, Goh, Gray, Voss, Radford, Chen,
  and Sutskever}]{ramesh2021zero}
Ramesh, A.; Pavlov, M.; Goh, G.; Gray, S.; Voss, C.; Radford, A.; Chen, M.; and
  Sutskever, I. 2021.
\newblock Zero-shot text-to-image generation.
\newblock In \emph{International Conference on Machine Learning}, 8821--8831.
  PMLR.

\bibitem[{Ravi and Larochelle(2017)}]{DBLP:conf/iclr/RaviL17}
Ravi, S.; and Larochelle, H. 2017.
\newblock Optimization as a Model for Few-Shot Learning.
\newblock In \emph{5th International Conference on Learning Representations,
  {ICLR} 2017, Toulon, France, April 24-26, 2017, Conference Track
  Proceedings}. OpenReview.net.

\bibitem[{Russakovsky et~al.(2015)Russakovsky, Deng, Su, Krause, Satheesh, Ma,
  Huang, Karpathy, Khosla, Bernstein et~al.}]{russakovsky2015imagenet}
Russakovsky, O.; Deng, J.; Su, H.; Krause, J.; Satheesh, S.; Ma, S.; Huang, Z.;
  Karpathy, A.; Khosla, A.; Bernstein, M.; et~al. 2015.
\newblock Imagenet large scale visual recognition challenge.
\newblock \emph{International journal of computer vision}, 115: 211--252.

\bibitem[{Schreiner(2023)}]{schreiner2023gpt4}
Schreiner, M. 2023.
\newblock GPT-4 architecture, datasets, costs and more leaked.
\newblock \emph{THE DECODER}.

\bibitem[{Sofiiuk, Petrov, and Konushin(2022)}]{sofiiuk2022reviving}
Sofiiuk, K.; Petrov, I.~A.; and Konushin, A. 2022.
\newblock Reviving iterative training with mask guidance for interactive
  segmentation.
\newblock In \emph{2022 IEEE International Conference on Image Processing
  (ICIP)}, 3141--3145. IEEE.

\bibitem[{Song et~al.(2023)Song, Yoon, Choi, and
  Avrithis}]{DBLP:conf/wacv/SongYCA23}
Song, C.~H.; Yoon, J.; Choi, S.; and Avrithis, Y. 2023.
\newblock Boosting vision transformers for image retrieval.
\newblock In \emph{{IEEE/CVF} Winter Conference on Applications of Computer
  Vision, {WACV} 2023, Waikoloa, HI, USA, January 2-7, 2023}, 107--117. {IEEE}.

\bibitem[{Sun et~al.(2023)Sun, Zhong, Fan, Tang, and
  Tai}]{DBLP:journals/UniBoost}
Sun, Y.; Zhong, Z.; Fan, Q.; Tang, C.; and Tai, Y. 2023.
\newblock UniBoost: Unsupervised Unimodal Pre-training for Boosting Zero-shot
  Vision-Language Tasks.
\newblock \emph{CoRR}, abs/2306.04715.

\bibitem[{Tao et~al.(2022)Tao, Zhu, Huang, Qiao, Wang, and
  Dai}]{tao2022siamese}
Tao, C.; Zhu, X.; Huang, G.; Qiao, Y.; Wang, X.; and Dai, J. 2022.
\newblock Siamese image modeling for self-supervised vision representation
  learning.
\newblock \emph{arXiv preprint arXiv:2206.01204}.

\bibitem[{Vaswani et~al.(2017)Vaswani, Shazeer, Parmar, Uszkoreit, Jones,
  Gomez, Kaiser, and Polosukhin}]{vaswani2017attention}
Vaswani, A.; Shazeer, N.; Parmar, N.; Uszkoreit, J.; Jones, L.; Gomez, A.~N.;
  Kaiser, {\L}.; and Polosukhin, I. 2017.
\newblock Attention is all you need.
\newblock \emph{Advances in neural information processing systems}, 30.

\bibitem[{Wei et~al.(2021)Wei, Fan, Xie, Wu, Yuille, and
  Feichtenhofer}]{wei_masked_2021}
Wei, C.; Fan, H.; Xie, S.; Wu, C.-Y.; Yuille, A.; and Feichtenhofer, C. 2021.
\newblock Masked Feature Prediction for Self-Supervised Visual Pre-Training.
\newblock \emph{ArXiv preprint}, abs/2112.09133.

\bibitem[{Wu and Mo(2022)}]{wu2022objectwise}
Wu, J.; and Mo, S. 2022.
\newblock Object-wise Masked Autoencoders for Fast Pre-training.
\newblock \emph{arXiv preprint arXiv:2205.14338}.

\bibitem[{Xie et~al.(2022)Xie, Zhang, Cao, Lin, Bao, Yao, Dai, and
  Hu}]{xie_simmim_2022}
Xie, Z.; Zhang, Z.; Cao, Y.; Lin, Y.; Bao, J.; Yao, Z.; Dai, Q.; and Hu, H.
  2022.
\newblock {SimMIM}: a Simple Framework for Masked Image Modeling.
\newblock In \emph{2022 {IEEE}/{CVF} Conference on Computer Vision and Pattern
  Recognition ({CVPR})}, 9643--9653. {IEEE}.
\newblock ISBN 978-1-66546-946-3.

\bibitem[{Yao, Desai, and Palaniswami(2022)}]{yao2022masked}
Yao, Y.; Desai, N.; and Palaniswami, M. 2022.
\newblock Masked Contrastive Representation Learning.
\newblock \emph{arXiv preprint arXiv:2211.06012}.

\bibitem[{Zbontar et~al.(2021)Zbontar, Jing, Misra, {LeCun}, and
  Deny}]{zbontar_barlow_2021}
Zbontar, J.; Jing, L.; Misra, I.; {LeCun}, Y.; and Deny, S. 2021.
\newblock Barlow Twins: Self-Supervised Learning via Redundancy Reduction.
\newblock \emph{ArXiv preprint}, abs/2103.03230.

\bibitem[{Zhou et~al.(2019)Zhou, Zhao, Puig, Xiao, Fidler, Barriuso, and
  Torralba}]{zhou2019semanticADE}
Zhou, B.; Zhao, H.; Puig, X.; Xiao, T.; Fidler, S.; Barriuso, A.; and Torralba,
  A. 2019.
\newblock Semantic understanding of scenes through the ade20k dataset.
\newblock \emph{International Journal of Computer Vision}, 127: 302--321.

\bibitem[{Zhou et~al.(2021)Zhou, Wei, Wang, Shen, Xie, Yuille, and
  Kong}]{zhou_ibot_2022}
Zhou, J.; Wei, C.; Wang, H.; Shen, W.; Xie, C.; Yuille, A.; and Kong, T. 2021.
\newblock {iBOT}: Image {BERT} Pre-Training with Online Tokenizer.
\newblock \emph{ArXiv preprint}, abs/2111.07832.

\end{thebibliography}

\clearpage


\section{Supplementary Materials}

\subsection{Augmentation}\label{sec:imp}
\begin{figure}[!htb]
    \centering
    \includegraphics[width=.8\linewidth]{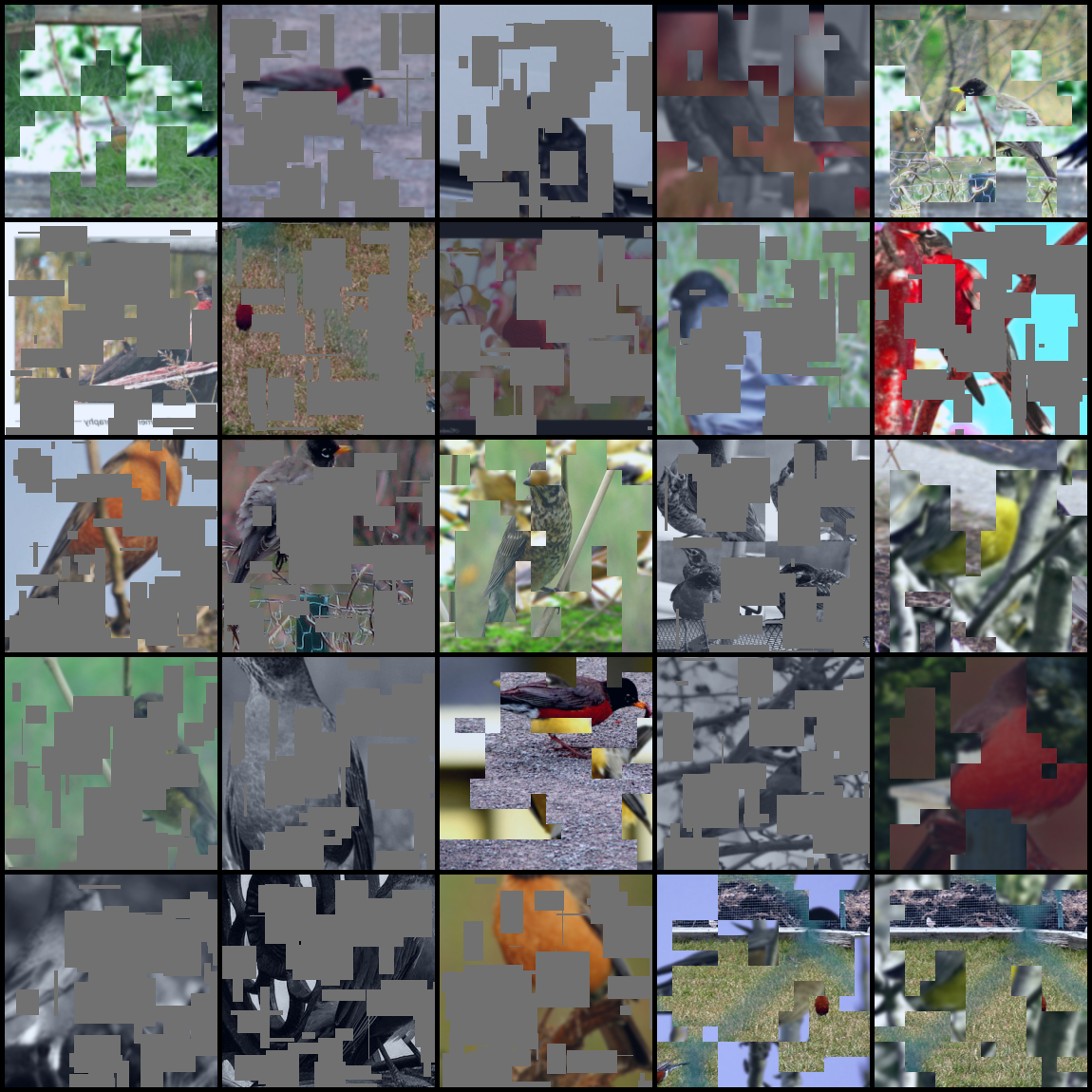}
    \caption{Masked views for pertaining. The images have a 35\% chance of being masked with other images.}
    \label{fig:aug}
\end{figure}

The augmentations include color jittering with parameters (0.4, 0.4, 0.4), Gaussian blur with a strength of 1.0, and solarization with a probability of 0.2, resulting in $\vx \in \mathbb{R}^{3 \times 224 \times 224}$. Inspired by the GMML~\cite{atito_gmml_2022,atito_sit_2021}, we utilize group-masked modeling to create the masked view $\vx^\mathtt{m}$. GMML involves employing a block-wise masking algorithm, where we selectively mask certain blocks of the input image. Specifically, we randomly replace 75\% of the masked blocks with noise or replace 30\% of the masked blocks from another image. The samples of augmented images are illustrated in Figure~\ref{fig:aug}.

\subsection{Evaluation Protocols}\label{sec:protocols}

\paragraph{Head Type.}
In general, methods based on global invariance tend to utilize the CLS token, which is paired with a linear head to perform classification tasks. On the other hand, methods grounded in local invariance commonly use the mean of patch tokens for their operations. The choice of ``head type''—that is, the mechanism through which the model processes its input data—is of substantial importance in the context of SSL (SSL) methods, as it can significantly influence the final performance of the model.
To put it another way, the performance observed when different head types are employed can reveal insights into where the semantic information is located or represented within the model.

In this study, we investigate three distinct head types, each using a different form of input data:
\begin{itemize}
    \item The CLS token alone (referred to as 'CLS').
    \item The mean of the patch tokens (referred to as 'PAT')
    \item A concatenation of the CLS token and the mean of the patch tokens (referred to as 'Hyber')
\end{itemize}

These distinctions in head types allow us to scrutinize the ways in which different approaches to processing input data can influence the performance of a model, and thereby shed light on the effective positioning of semantic information within these various configurations.

\paragraph{$k$-NN.}
The \(k\)-Nearest Neighbors (\(k\)NN) algorithm is a well-regarded technique commonly employed for classification and regression tasks. It is categorized as a non-parametric and lazy learning algorithm. Being non-parametric means that \(k\)NN does not make any underlying assumptions about the distribution of the data. As a lazy learner, it does not involve a training phase to construct a generalized model. Therefore, in the context of unsupervised learning, \(k\)NN serves as a valuable tool for evaluating the quality of learned representations. This is because it can effectively measure how well the learned feature space clusters or separates different classes based on a simple, intuitive principle—namely, that similar instances are near to each other in this space.

In our study, as documented in Table~\ref{tab:knn_setting}, we examine the impact of two key parameters: the number of nearest neighbours (\(k\)) and the head type. From our analysis, it is evident that our proposed model (MMC) yields the most favorable results when \(k\) is set to 10 and the mean of patch tokens is utilized as the head type. 
This finding suggests that, within the context of the MMC method, semantic information is principally promoted and emphasized in local feature representations. In other words, the MMC method appears to be particularly adept at capturing and leveraging the nuanced, localized information inherent in individual data patches, rather than in more global or holistic data features. Therefore, we employ $k=10$ in our experiments.

\begin{table}[htb]
\centering

\begin{tabular}{llccc}
\toprule
   Head                &  K   & Pets          & Flowers       & CIFAR10       \\
\midrule
\multirow{4}{*}{CLS}   & 10  & \textbf{72.9} & \textbf{63.5} & \textbf{84.5} \\
                       & 20  & 73.4          & 59.6          & 83.9          \\
                       & 50  & 71.6          & 54.8          & 82.7          \\
                       & 100 & 70.7          & 52.9          & 80.9          \\
                       \midrule
\multirow{4}{*}{PAT}   & 10  & \textbf{75.9} & \textbf{66.0} & \textbf{84.9} \\
                       & 20  & 76.4          & 63.8          & 84.5          \\
                       & 50  & 75.2          & 62.0          & 83.3          \\
                       & 100 & 75.0          & 61.1          & 82.0          \\
                       \midrule
\multirow{4}{*}{Hyber} & 10  & 75.0          & \textbf{65.5} & \textbf{84.8} \\
                       & 20  & \textbf{75.3} & 62.5          & 84.6          \\
                       & 50  & 74.0          & 59.4          & 83.0          \\
                       & 100 & 73.4          & 58.2          & 81.8    \\
                       \bottomrule
\end{tabular}
\caption{Comparison of Classification Performance across Different Head Types and \(k\). Bold values indicate the highest accuracy achieved for each head type and dataset combination.}

\label{tab:knn_setting}
\end{table}

\paragraph{Finetune.}
Finetuning is a prevalent step in transfer learning, which involves adapting a model pre-trained on a large dataset to a new, typically smaller, dataset that may belong to a similar task or domain. To strike a balance between preserving the knowledge learned from the source domain and effectively adapting the model to the new target domain, the technique of ``layer decay'' can be employed. Layer decay involves applying different, gradually diminishing learning rates to various layers of the neural network. This strategy allows for more subtle adjustments in the earlier layers. Our recipe is depicted in Table~\ref{tab:finetune_recipe}.

\begin{table}[!htb]
    \centering
    \setlength{\tabcolsep}{5pt}
\begin{tabular}{lp{2.5cm}p{2.5cm}}
\toprule
                       & IN1K              & Others            \\
\midrule
optimizer              & AdamW             & AdamW             \\
weight decay           & 0.05                 & 0.05                \\
layer decay            & 0.65              & 0.65 \\
momentum     & 0.9               & 0.9               \\
lr    & 5e-4             & 5e-4             \\
lr schedule & cosine decay      & cosine decay      \\
\midrule
batch size             & 64               & 64               \\
warmup           & 5                 & 5                 \\
epochs        & 100                & 200                \\
augmentation           & \small rand-m9-mstd0.5-inc1 & \small ThreeAugment\\
GPUs                   & 4                 & 2                \\
\bottomrule
\end{tabular}
    \caption{Finetuning recipe for transfer learning}
    \label{tab:finetune_recipe}
\end{table}

\paragraph{Segmentation.}
Semantic segmentation can be interpreted as a classification problem conducted at the level of individual pixels. To prepare the images for processing, they are resized and cropped to a 512x512 resolution, and then augmented using a Random Flip and Photometric Distortion techniques. The optimization algorithm used in this process is AdamW, with a learning rate (lr) of 0.0001, betas set at (0.9, 0.999), and a weight decay of 0.05. 
In terms of learning rate scheduling, a linear policy is employed with a warm-up phase, consisting of 1500 iterations. This approach mirrors the linear evaluation protocol used in classification tasks. For evaluation purposes, the frozen features from layers 5, 7, 9, and 11 are analyzed, and only a linear layer is fine-tuned. This approach allows for a clearer comparison of the quality of the different representations generated by the model.
All models are trained using 4 GPUs for a total of 40,000 iterations. During this training process, each GPU processes 4 samples. Importantly, this configuration remains consistent across all the datasets employed in this study, which include ADE20K, Cityscapes, and VOC augmented datasets.

\subsection{Additonal Result}

\paragraph{Classification.}
Our framework can also deal with classification tasks via the CLS token. We treat the CLS token $\vz_{\mathtt{CLS},s}$ of a support image as a prompt for classification. A query image is an input to predict its class. We calculate the cosine similarity for the query and the prompt $\sigma(\vz_{\mathtt{CLS},q},\vz_{\mathtt{CLS},s})$. 
Our algorithm predicts whether $\vx_q$ and $\vx_s$ have the same class based on whether the similarity is over a threshold $T$ or not.
This setting is helpful when dealing with open-wide problems where there are agnostic negative classes. In this setting, our algorithm performs binary classification, therefore, we utilize f1 score to measure the performance.

We rigorously evaluate the efficacy of the CLS token through the employment of both the nearest neighbor algorithm and our proposed threshold-based algorithm on three datasets: miniImageNet, Pets, and CIFAR10.
The mini-ImageNet (miniIN) is a variant of ImageNet-1K that consists of 100 classes with 600 image samples per each, which's images are seen. In our implementation, we re-scale the original image from 84×84 to 224x224 to fit our model and use the 20 testing categories split by ~\citeauthor{DBLP:conf/iclr/RaviL17}. Pets~\cite{parkhi2012cats} contains 37 categories with roughly 200 images for each class, which is very challenging as these animals, particularly cats, are very deformable and the difference can be quite subtle. CIFAR10~\cite{krizhevsky2009learning_cifar10} consists of 60,000 32x32 color images in 10 different classes, which are distinguishable from each other.
We consider the 5-way 1-shot setting. 
Our reporting methodology involves sweeping the threshold across the range of 0.1 to 0.9 in increments of 0.1 and shows the highest f1 score for each experiment. As discerned from Table~\ref{tab:fewshot_classification}, it becomes evident that our model trails behind iBOT on miniIN and CIFAR10 but outperforms it on Pets. This observation underscores the strengths of DINO and iBOT in identifying similar objects, while emphasizing the potential downsides when dealing with distinct and subtle differences in the case of iBOT.

\begin{table}[t]
\centering
\setlength{\tabcolsep}{5pt}
\scalebox{0.96}{
\begin{tabular}{lcccccc@{}}
\toprule
        & \multicolumn{3}{c}{Accuracy}                  & \multicolumn{3}{c}{F1}                        \\ 
        & \small miniIN        & \small CF10       & \small Pets          &  \small  miniIN        &\small  CF10       & \small Pets         \\ 
          \cmidrule(lr){2-4} \cmidrule(lr){5-7}
SiT     & {72.6} & \textbf{50.9} & 80.9          & {57.3} & \textbf{45.4} & 67.8          \\
DINO    & 73.1          & 40.9          & 83.7          & 65.0          & 39.2          & 73.0          \\
iBOT    & \textbf{74.1} & 39.9          & \textbf{84.6} & \textbf{66.0} & 37.7          & \textbf{73.4} \\
MAE     & 27.1          & 27.2          & 24.4          & 33.8          & 34.7          & 33.5          \\
MC-SSL & 60.0          & 30.8          & 80.8          & 50.3          & 33.7          & 70.1          \\
MMC (ours)     & 73.3          & 47.4          & 69.7          & 62.3          & 42.9          & 62.1          \\ 
\bottomrule
\end{tabular}
}
\caption{Performance comparison for 5-way 1-shot on miniImageNet (miniIN), Pets, and CIFAR10 (CF10). Accuracy is calculated by the nearest neighbour. F1 refers to the best harmonic mean of the precision and recall with an optimal threshold.}
\label{tab:fewshot_classification}
\end{table}

\paragraph{Image Retrieval.}
Image retrieval involves searching and retrieving relevant images from a large collection of images for a given query image~\cite{DBLP:journals/corr/IRT}. We take the CLS token and directly apply \(k\)-NN for retrieval. We report the Mean Average Precision (mAP) for the Medium (M) and Hard (H) splits on the revisited Oxford and Paris image retrieval datasets~\cite{DBLP:conf/cvpr/RadenovicITAC18}. In Table~\ref{tab:ir}, we present a comparison between two different types of image retrieval models: 
1) An off-the-shelf feature extraction approach using a SSL (SSL) Vision Transformer-B (ViT-B) model pre-trained on the ImageNet1K dataset, and 
2) State-of-the-art (SOTA) image retrieval models that were trained with supervision. 
In our comparison, IRT\_O and IRT\_R ~\cite{DBLP:journals/corr/IRT} are models that were trained with class-wise pretraining using supervision and contrastive methods, respectively. DToP~\cite{DBLP:conf/wacv/SongYCA23}, on the other hand, represents the current state-of-the-art model for image retrieval tasks.
From our observations, it is notable that the SSL models outperform the ViT that was pretrained with supervision (IRT\_O). This suggests that SSL is particularly effective at extracting meaningful image representations without requiring label information. 
While there remains a significant performance gap between our SSL approach and the current state-of-the-art (SOTA) model, DToP, it is important to consider the differences in architectural design and datasets used between these approaches.
It is noteworthy that IRT\_R requires class labels for its class-wise contrastive training. Despite this, our SSL models demonstrate competitive performance compared to IRT\_R, highlighting the potential of SSL algorithms for effective image retrieval tasks in an entirely unsupervised manner.

\begin{table}[]
\centering
\begin{tabular}{@{}l@{\hspace{0pt}}c@{\hspace{3pt}}ccccc@{}}
\toprule
Model   & Arch & Data & \multicolumn{2}{c}{ROxford} & \multicolumn{2}{c}{RParis} \\                
        &      &      & M          & H          & M           & H          \\ 
\cmidrule(lr){4-5} \cmidrule(lr){6-7}
IRT\_O  & ViT-B & IN1K & 20.3       & 5.6        & 50.2        & 26.3       \\
IRT\_R  & ViT-B & IN1K & 34.0       & 11.5       & 66.1        & 40.2       \\ 
DToP    & ViT-B+R50& SFM & 64.8	  & 41.3       & 82	        & 63.2       \\
\midrule
iBOT    & ViT-B & IN1K & \textbf{34.2} & \textbf{13.4} & \textbf{61.9} & \textbf{35.2} \\
DINO    & ViT-B & IN1K & 32.8       & 10.7       & 60.4        & 34.2       \\
MC\_SSL & ViT-B & IN1K & 27.4       & 11.4       & 52.6        & 29.9       \\
SiT     & ViT-B & IN1K & 29.3       & 8.3        & 51.1        & 21.3       \\
MMC (ours)     & ViT-B & IN1K & 26.1       & 8.9        & 53.4        & 28.6       \\
\bottomrule
\end{tabular}
\caption{Image Retrieval on ROxford and RParis. The methods in the upper part utilize supervision. R50 denotes ResNet50.
}\label{tab:ir}
\end{table}

\subsection{Property of MMC}

\begin{figure}
    \centering
    \includegraphics[width=\linewidth]{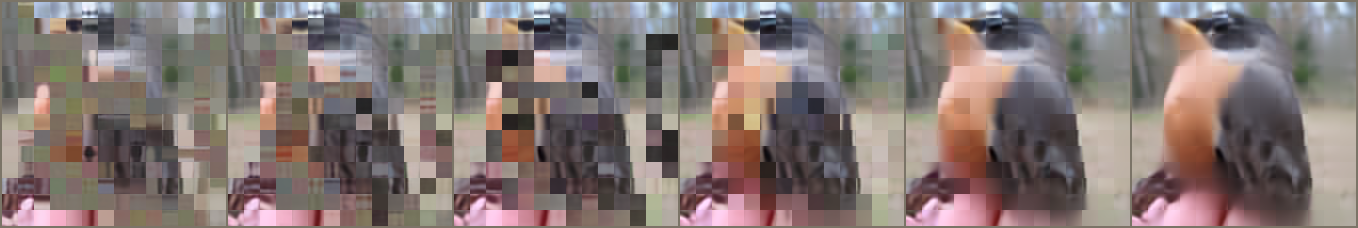}
    \includegraphics[width=\linewidth]{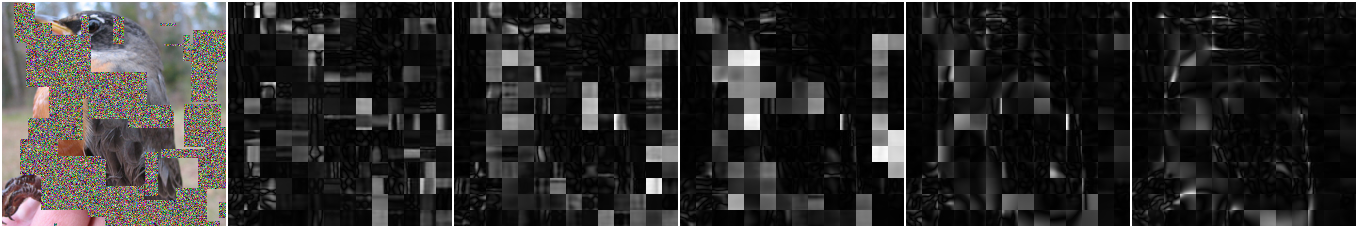}
    \caption{Reconstruction visualization of MMC layer by layer. The first row shows the reconstruction from layers 7 to 12. The first picture in the second row is the input image, and the rests are the difference map of reconstruction between two adjacent layers. }
    \label{fig:viz_layers}
\end{figure}

\paragraph{Reconstruction from Layers.}
In our analysis of MMC, we visualize the reconstruction process from layers 7 to 12 for a given masked image. To achieve this, we take the output of a specific layer and feed it to a reconstruction head, bypassing the subsequent layers in the ViT. Through this approach, we are able to gain insights into how each layer of the transformer network contributes to the task of image reconstruction. 
As illustrated in Figure~\ref{fig:viz_layers}, it is apparent that different layers of the transformer are responsible for recovering different regions of the masked image. We also present difference maps of the reconstructions between two adjacent layers to provide a clearer picture of the roles that various layers play in the reconstruction process.
Our findings indicate that layer 10 predominantly focuses on the reconstruction of the masked portions of the image and the subsequent layers 11 and 12 refine the details to generate a coherent and visually pleasing reconstruction. This suggests a function differentiation in ViTs.

\paragraph{Feature Variance.}\label{sec:perturbation}
In our study, we examine the stability of the representations learned through self-supervised methods under various forms of disturbance. We evaluate this stability under two key concepts: image-level and patch-level consistency. For the image-level concept, our hypothesis is that the CLS token, which is obtained from the same image but under different global views, should exhibit similarity. That is, despite changes in the global view of the image, the high-level representation captured by the CLS token should remain relatively consistent.
For the patch-level concept, we posit that the patch tokens extracted from the same image, but with different masked areas, should also be similar. This concept examines the stability of representations at a more granular, localized scale within the image. 
To evaluate these concepts rigorously, we conduct experiments to assess the representations' consistency when subjected to various perturbations. In our experiments, we apply two typical perturbations used in SSL (SSL): RandomCrop for image-level consistency and block-wise masking for patch-level consistency. 
Specifically, for each image in IN1K test set, we generate 128 different views (i.e., variations achieved through random cropping) and compute their corresponding image representations from the outputs of the last block of our model. We then calculate the variance of these representations for a given image, with the expectation that a lower variance indicates more stable and consistent representations across different views.
Similarly, to evaluate patch-level consistency, we randomly mask 50\% of an image with noise and generate 128 masked views for each image. After applying this masking, we compare the outputs of the last block in the same position across these differently-masked versions of the image.
As presented in Table~\ref{tab:rep_stab}, our results indicate that our method achieves a low level of variance across different disturbances, suggesting that it produces highly confident and stable representations. 

\begin{table}[!hbt]
\centering
\caption{The average variance of the CLS token for global views and patch tokens for masked views. Low variance means the representation of one image is consistent in capturing concepts. }\label{tab:rep_stab}
\scalebox{0.9}{
\begin{tabular}{@{}lccccc@{}}
\toprule
 & MAE & DINO & iBOT & MSN  & MMC \\ 
 \midrule
$\mathtt{CLS}$       & \textbf{0.12}     & 0.43          & 0.33          & 0.34                       & 0.19          \\
$\mathtt{PAT}$     & 0.55         & 1.00          & 0.75          & 0.56                      & \textbf{0.32}          \\ 
\bottomrule
\end{tabular}
}
\end{table}

\paragraph{Patch Semantics vs. CLS Semantics.}
In our study, we compared the classification performance of both global features (CLS token) and local features (patch tokens) using the \(k\)-Nearest Neighbors (k-NN) protocol. The results of this comparison are presented in Table~\ref{tab:semantics}. One of our key observations from this comparison is the distinct performance characteristics between the CLS token and patch tokens across different SSL methods.
Specifically, while the CLS token consistently yields high performance, we observe a notable decline in performance when using patch tokens for the iBOT and DINO methods. This significant diminishment in performance for patch tokens suggests that these SSL methods struggle to effectively leverage and transfer semantic information at a more granular, patch-level scale. Essentially, they appear to excel at capturing high-level, global representations of images (as evidenced by the strong performance of their CLS token), but this prowess does not extend to the more detailed, localized patch tokens.

In contrast, our proposed model MMC exhibits a more robust ability to transfer semantic information from the CLS token to the patch tokens. This is evident from the relatively stable and high performance of our model, regardless of whether we are considering CLS or patch tokens. This suggests that our model is not only adept at capturing global, high-level semantic information about images but is also effective at maintaining this semantic richness when we examine the image at a finer, patch-level scale. 

\begin{table}[]
\centering
\begin{tabular}{lcc}
\toprule
     & PAT & CLS \\
\midrule
MAE & 35.32 & 46.98 \\
DINO & 75.56 & 91.54 \\
iBOT & 75.22 & \textbf{92.12} \\
MSN & 82.38 & 89.38 \\
SiT & 76.40 & 87.62 \\
MC-SSL & 83.38 & 88.98 \\
MMC (ours) & \textbf{90.04} & 89.94 \\
\bottomrule
\end{tabular}
\caption{$k$-NN performance for patch tokens (PAT) and the CLS token (CLS) on IN100.}\label{tab:semantics}
\end{table}

\end{document}